\documentclass[10pt,twocolumn,letterpaper]{article}

\usepackage[pagenumbers]{cvpr} 

\usepackage{graphicx}
\usepackage{amsmath}
\usepackage{amssymb}
\usepackage{booktabs}
\usepackage{textcomp}  
\usepackage{gensymb}  
\usepackage{siunitx}

\usepackage{stmaryrd}   
\usepackage{trimclip}   

\usepackage{pifont}   
\usepackage{xcolor}    
\usepackage{colortbl}

\usepackage{makecell}  
\usepackage[flushleft]{threeparttable}  
\usepackage{multirow}
\usepackage{multicol}
\usepackage{enumitem}

\usepackage[pagebackref,breaklinks,colorlinks]{hyperref}

\usepackage[capitalize]{cleveref}
\crefname{section}{Sec.}{Secs.}
\Crefname{section}{Section}{Sections}
\Crefname{table}{Table}{Tables}
\crefname{table}{Tab.}{Tabs.}

\def\cvprPaperID{31} 
\def\confName{CVPR}
\def\confYear{2023}

\DeclareMathOperator*{\argmax}{arg\,max}

\newcommand{\cmark}{\ding{51}}  

\definecolor{Gray}{gray}{0.9}

\makeatletter
\DeclareRobustCommand{\shortto}{%
  \mathrel{\mathpalette\short@to\relax}%
}
\newcommand{\short@to}[2]{%
  \mkern2mu
  \clipbox{{.25\width} 0 0 0}{$\m@th#1\vphantom{+}{\shortrightarrow}$}%
  }
\makeatother

\newcommand{\net}{CoVIO\xspace}

\begin{document}

\title{CoVIO: Online Continual Learning for Visual-Inertial Odometry}


\author{Niclas Vödisch$^{1}$
\hspace{1em} Daniele Cattaneo$^{1}$
\hspace{1em} Wolfram Burgard$^{2}$
\hspace{1em} Abhinav Valada$^{1}$ \vspace{0.1cm} \\
$^1$University of Freiburg 
\hspace{0.65em} $^2$University of Technology Nuremberg}

\maketitle

\begin{abstract}
    Visual odometry is a fundamental task for many applications on mobile devices and robotic platforms. Since such applications are oftentimes not limited to predefined target domains and learning-based vision systems are known to generalize poorly to unseen environments, methods for continual adaptation during inference time are of significant interest.
In this work, we introduce \net for online continual learning of visual-inertial odometry. \net effectively adapts to new domains while mitigating catastrophic forgetting by exploiting experience replay. In particular, we propose a novel sampling strategy to maximize image diversity in a fixed-size replay buffer that targets the limited storage capacity of embedded devices. We further provide an asynchronous version that decouples the odometry estimation from the network weight update step enabling continuous inference in real time.
We extensively evaluate \net on various real-world datasets demonstrating that it successfully adapts to new domains while outperforming previous methods.
The code of our work is publicly available at \mbox{\small \url{http://continual-slam.cs.uni-freiburg.de}}.

\end{abstract}

\section{Introduction}
\label{sec:introduction}

Reliable estimation of a robot's motion based on its onboard sensors is a fundamental requirement for many downstream tasks including localization and navigation. Devices such as inertial measurement units (IMU) or inertial navigation systems (INS) provide a way to directly measure the robot's motion based on acceleration and GNSS readings. An alternative is to use visual odometry (VO) leveraging image data from monocular or stereo cameras. Such VO methods have been successfully used in UAVs~\cite{fu2015efficient}, mobile applications~\cite{schoeps2014semidense}, and even mars rovers~\cite{maimone2007two}. Similar to other vision tasks, learning-based VO has gained increasing attention as the learnable high-level features can circumvent problems in textureless regions~\cite{valada2018incorporating, valada2018deep} or in the presence of dynamic objects~\cite{bevsic2022dynamic} where classical handcrafted methods suffer. However, learning-based VO lacks the ability to generalize to unseen domains, hindering their open-world deployment. Recently, adaptive VO~\cite{luo2019real} has opened a new avenue of research, \eg, by using continual learning (CL) methodologies to enhance VO during inference time~\cite{voedisch2023continual}.

\begin{figure}[t]
    \centering
    \includegraphics[width=\linewidth]{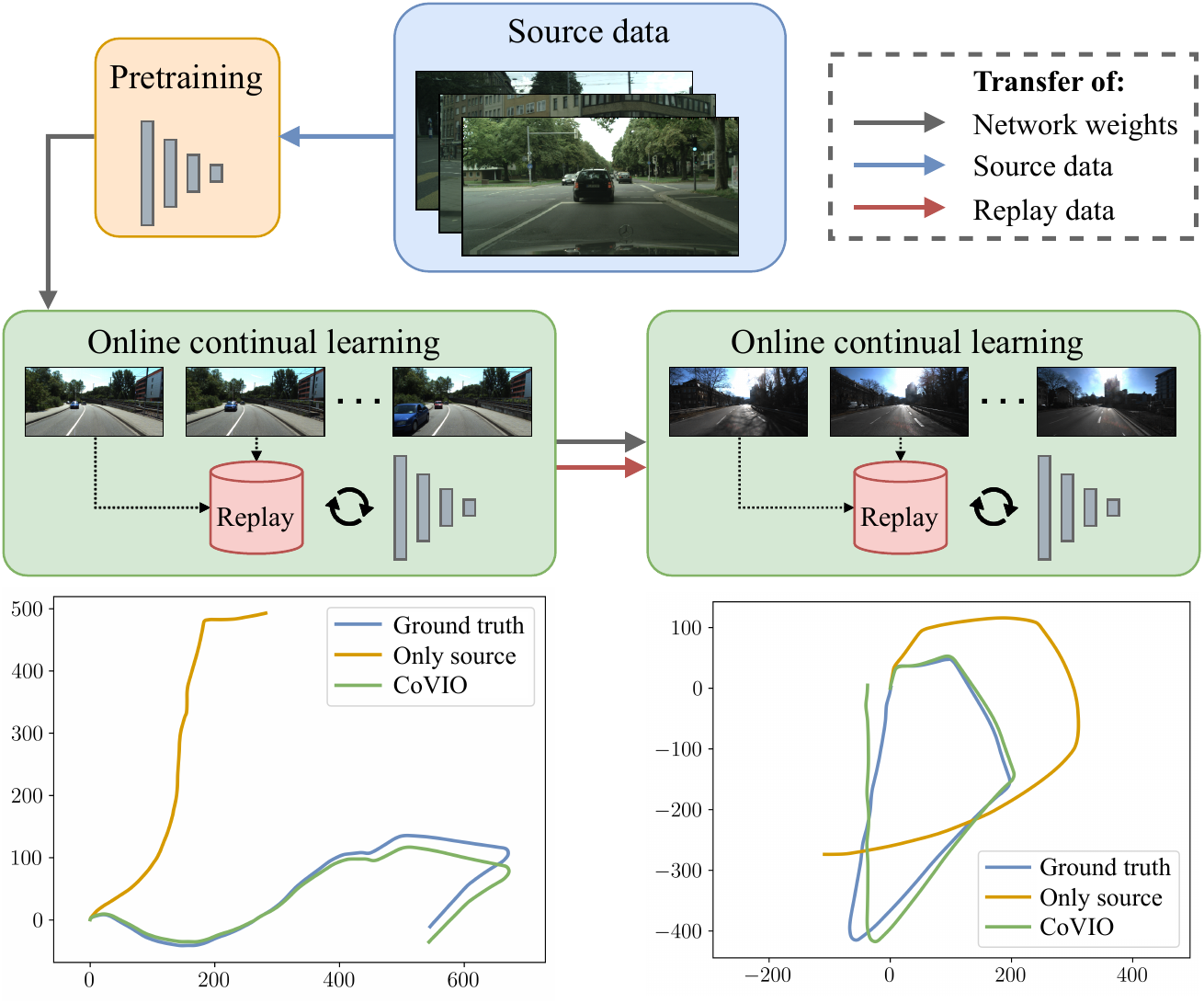}
    \caption{We propose \net for online continual learning of visual-inertial odometry. After pretraining on a source domain that is then discarded, \net further updates the network weights during inference on a target domain. Using experience replay \net successfully mitigates catastrophic forgetting.}
    \label{fig:teaser}
\end{figure}

Most commonly, learning-based VO leverages monocular depth estimation as an auxiliary task~\cite{luo2019real, li2020self, voedisch2023continual} and exploits an unsupervised joint training scheme of a PoseNet, estimating the camera motion between two frames, and a DepthNet, estimating depth from a single image~\cite{godard2019digging}. Due to the unsupervised nature of this approach, learning-based VO can be continuously trained also during inference time.
In addition to classical domain adaptation~\cite{besic2022unsupervised}, where knowledge is transferred from a single source to a single target domain, the recent study on continual SLAM~\cite{voedisch2023continual} also investigates a sequential multi-domain setting as illustrated in \cref{fig:teaser}. The authors introduce CL-SLAM, which fuses adaptive VO with a graph-based SLAM backend.
To avoid catastrophic forgetting, \ie, overfitting to the current domain while losing the ability to perform well on past domains, CL-SLAM employs a dual-network architecture comprising an expert and a generalizer for both efficient domain adaptation and knowledge retention combined with experience replay.
However, the previously proposed CL-SLAM suffers from three main drawbacks:
First, network weights are transferred from the generalizer to the expert upon the start of a new evaluation sequence, \ie, a human supervisor decides when new data should be classified as a domain change.
Second, the utilized replay buffer of the generalizer is of infinite size and, thus, does not consider the limited storage capacity of real-world applications.
Finally, since every received frame triggers an update of the network weights before yielding the VO estimate, real-time usage is difficult to achieve on low-power devices such as embedded hardware in robots.

In this work, we propose a novel adaptive visual-inertial odometry estimation method called \net that explicitly addresses all of the aforementioned drawbacks of \mbox{CL-SLAM}. Similar to Kuznietsov~\etal~\cite{kuznietsov2022towards}, we consider a source-free setting, \ie, experience replay does not include data from the source domain used for pretraining. 
In particular, the contributions of this work can be summarized as follows:
\begin{enumerate}[topsep=0pt, noitemsep]
    \item We replace the dual-network architecture with a single network addressing both domain adaptation and knowledge retention but simplifying the overall architecture and reducing the GPU memory footprint. Additionally, this resolves the issue of transferring network weights without domain classification.
    \item We propose a fixed-size replay buffer that maximizes image diversity and addresses the limited storage capacity of embedded devices.
    \item We present an asynchronous version of \net that separates the core motion estimation from the network update step allowing true continuous inference
    \item We perform extensive evaluations of \net on various datasets, both publicly available and in-house, demonstrating its efficacy compared to other visual odometry methods.
    \item We release the code of our work and trained models at \mbox{\small \url{http://continual-slam.cs.uni-freiburg.de}}.
\end{enumerate}

\section{Related Work}

In this section, we provide a brief introduction to continual and lifelong learning and summarize previous methods for domain adaptation of learning-based visual odometry.


{\parskip=5pt
\noindent\textit{Continual Learning:}
Deep learning-based models are commonly trained for a specific task, which is defined a priori, using a fixed set of training data. During inference, the model is then employed on previously unseen data from the same domain without further updates of the network weights. However, in many real-world scenarios, this assumption does not hold true, \eg, the initially used training data might not well represent the data seen during inference, thus leading to a domain gap and suboptimal performance. Additionally, the objective of the task can change over time.
Continual learning (CL) and lifelong learning~\cite{thrun1995is} aim to overcome these challenges by enabling a method to continually learn additional tasks given new training data. In contrast to vanilla domain adaptation~\cite{besic2022unsupervised}, CL methods should maintain the capability to solve previously learned tasks, \ie, avoiding catastrophic forgetting. Ideally, learning a task also yields improved performance on previous tasks (\textit{positive backward transfer}) as well as on yet unknown future tasks (\textit{positive forward transfer})~\cite{lopez2017gradient}.
The majority of CL approaches can be categorized into three strategies. First, experience replay directly tackles catastrophic forgetting from a data-driven perspective. For instance, both CoMoDA~\cite{kuznietsov2021comoda} and CL-SLAM~\cite{voedisch2023continual} store images in a replay buffer and combine online data with replay samples when updating the network weights. Second, regularization techniques such as knowledge distillation~\cite{valverde2021there} preserve information on a more abstract feature level. Finally, architectural methods prevent forgetting by using certain network structures, \eg, LSTMs~\cite{li2020self} and dual-network architectures~\cite{voedisch2023continual}, or by directly freezing internal model parameters.
Online continual learning~\cite{lunayach2022lifelong, wang2021wanderlust} describes an extension of CL by considering a setting, where the model is continuously updated on a stream of data during inference time. Online CL also includes scenarios, which gradually change from one domain to another~\cite{taufique2022unsupervised}.
In this work, we employ online CL with experience replay for learning-based visual-inertial odometry estimation.
}


\begin{figure*}[t]
    \centering
    \includegraphics[width=\linewidth]{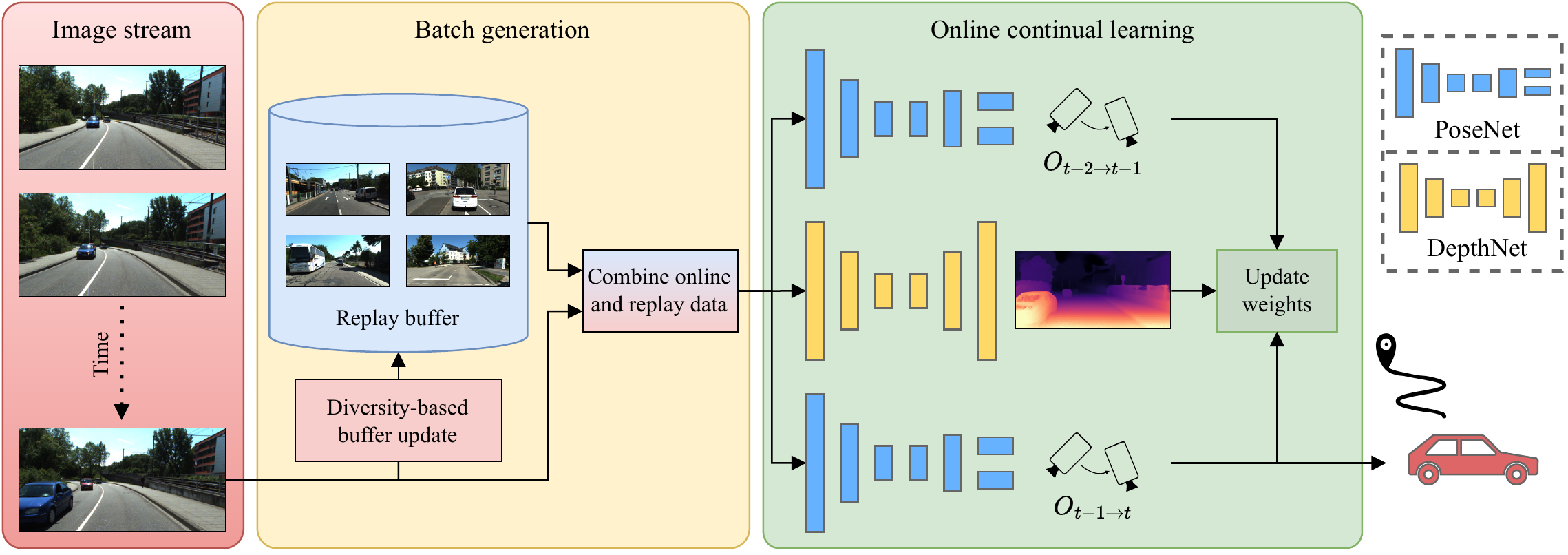}
    \caption{Our proposed \net performs online continual learning on a stream of RGB images leveraging unsupervised monocular depth estimation as an auxiliary task. In each update step, the image triplet consisting of the current and the two previous frames is combined with samples from a replay buffer and then fed to the networks to update their weights via backpropagation. The estimated camera motion between the previous and the current image corresponds to the generated VO output. The replay buffer is optionally updated if the current frame is sufficiently different from the existing content.}
    \label{fig:overview}
\end{figure*}

{\parskip=5pt
\noindent\textit{Adaptive Visual Odometry:}
Online adaptation of learning-based visual odometry (VO) and simultaneous localization and mapping (SLAM) aims to enhance performance on the fly allowing robotic systems to operate more reliably in previously unseen environments.
Most commonly, learning-based VO relies on monocular depth estimation as an auxiliary task enabling joint training of a DepthNet and a PoseNet in an unsupervised manner~\cite{godard2019digging}.
In an early work on adaptive VO,  Luo~\etal~\cite{luo2019real} accumulate images from an online camera stream and leverage the unsupervised training scheme to update both networks. Different to experience replay in CL, the buffer of accumulated images is emptied after the update step, \ie, each sample is only seen once.
Li~\etal~\cite{li2020self} propose an architectural CL technique, replacing the standard convolutional layers with LSTM variants to prevent forgetting. During deployment, the networks are continuously trained using only the online data. In a follow-up work by the same authors~\cite{li2021generalizing}, the PoseNet is substituted with optical flow-based point matching.
Similarly, GeoRefine~\cite{ji2022georefine} combines online depth refinement with dense visual mapping. While the DepthNet is updated following the aforementioned works, GeoRefine uses a non-adaptive odometry and tracking module based on optical flow.
Loo~\etal~\cite{loo2021online} propose an adaptive visual SLAM system that combines experience replay with a variant of elastic weight consolidation (EWC) to further regularize the weight updates of both the DepthNet and the PoseNet.
Finally, to avoid catastrophic forgetting in a multi-domain adaptation setting, \mbox{CL-SLAM}~\cite{voedisch2023continual} exploits a dual-network architecture, which is composed of an expert to perform effective online adaptation to the new domain and a generalizer to retain previously acquired knowledge by leveraging experience replay.
In this work, we propose an adaptive method for visual-inertial odometry built on \mbox{CL-SLAM} that explicitly addresses its shortcomings as outlined in \cref{sec:introduction}.
}

\section{Technical Approach}

In the following sections, we first describe the network architecture along with the pretraining procedure on a source domain. Then, we introduce \net and provide detailed explanations of all contributions.


\subsection{Network Architecture and Pretraining}
\label{ssec:ta-network-pretraining}

In this section, we detail the network architecture of our proposed \net and the loss functions that we employ during the initial training phase.

{\parskip=3pt
\noindent\textit{Network Architecture:}
We build our network following the common scheme of unsupervised monocular depth estimation leveraging two separate networks that we refer to as DepthNet and PoseNet as depicted in \cref{fig:overview}. Similar to \mbox{CL-SLAM}~\cite{voedisch2023continual}, for an image triplet $\{ \mathbf{I}_{t-2}, \mathbf{I}_{t-1}, \mathbf{I}_t \}$ we use Monodepth2~\cite{godard2019digging} to jointly predict a dense depth map $\mathbf{D}_{t-1}$ of the center image and the camera motion with respect to both neighboring frames, \ie, $\mathbf{O}_{t-2 \shortto t-1}$ and $\mathbf{O}_{t-1 \shortto t}$. In \net, we then output the latter as the VO estimate. In particular, we use an implementation comprising two separate ResNet-18~\cite{he2016deep} encoders for the DepthNet and the PoseNet.
}


{\parskip=3pt
\noindent\textit{Source Domain Pretraining:}
To initialize \net, we perform unsupervised training on a source domain $\mathcal{S}$ in an offline manner. In detail, we exploit the photometric reprojection loss $\mathcal{L}_\mathit{pr}$ and the image smoothness loss $\mathcal{L}_\mathit{sm}$ to train the DepthNet and the PoseNet~\cite{godard2019digging}. We additionally supervise the PoseNet with scalar velocity readings from the vehicle's IMU~\cite{guizilini20203d}. The applied velocity supervision term $\mathcal{L}_\mathit{vel}$ enforces metric scale-aware odometry estimates. Thus, our total loss is composed of three terms:
\begin{equation}
    \label{eqn:total_loss}
    \mathcal{L} = \mathcal{L}_{pr} + \gamma \mathcal{L}_{sm} + \lambda \mathcal{L}_{vel},
\end{equation}
with weighting factors $\gamma$ and $\lambda$.
}


\subsection{Online Continual Learning}
\label{ssec:ta-continual-learning}

After pretraining on a source domain $\mathcal{S}$, we use \net to perform online continual learning on an unseen target domain $\mathcal{T}$. As illustrated in \cref{fig:overview}, each new RGB image triggers the following steps:
\begin{enumerate}[label={(\arabic*)}, topsep=0pt, noitemsep]
    \item Create a data triplet comprising the new frame $\mathbf{I}_t$ and the two previous frames $\mathbf{I}_{t-1}$ and $\mathbf{I}_{t-2}$ along with the corresponding IMU readings.
    \item Check whether this triplet should be added to the replay buffer using the proposed diversity-based update mechanism.
    \item Sample from the replay buffer and combine the samples with the previously generated data triplet.
    \item Estimate the depth map $\mathbf{D}_{t-1}$ and the camera motions $\mathbf{O}_{t-2 \shortto t-1}$ and $\mathbf{O}_{t-1 \shortto t}$.
    \item Compute the loss defined in \cref{eqn:total_loss} and update the network weights via backpropagation.
    \item Repeat steps (4) and (5) for $c$ iterations.
    \item Output $\mathbf{O}_{t-1 \shortto t}$ as the odometry estimate.
\end{enumerate}

{\parskip=3pt
\noindent
In the following, we provide more details on the proposed replay buffer and the online continual learning strategy of \net. Finally, we propose an asynchronous version of \net that separates the motion estimation from the network update step allowing continuous inference.
}


\subsubsection{Replay Buffer}
\begin{figure}
    \centering
    \captionsetup[subfigure]{justification=justified, font=small, width=1.25\linewidth}
    \subfloat[An image is added to the replay buffer if the cosine similarity to the most similar image in the buffer is below a threshold, \eg, $\theta_\mathit{th} = 0.95$. Here, the image will be added since $0.92 < \theta_\mathit{th}$.]{%
        \includegraphics[width=.8\linewidth]{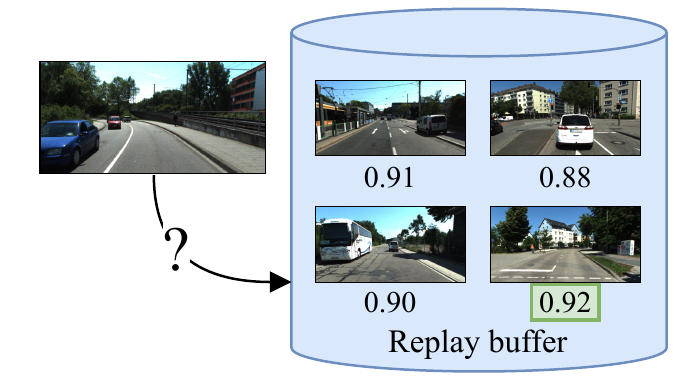}}
    \hfill
    \subfloat[If adding a new image results in exceeding the allowed size of the replay buffer, the image that is the most similar with respect to all other images is removed. The table shows the cosine distance between two frames.]{%
        \includegraphics[width=.8\linewidth]{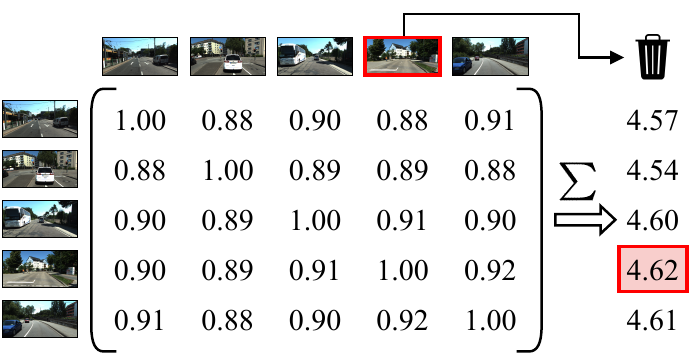}}
    \caption{Diversity-based update mechanism of the replay buffer, separated in (a) adding and (b) removing a sample.}
    \label{fig:replay-buffer}
\end{figure}

As outlined in \cref{sec:introduction}, previous works~\cite{kuznietsov2021comoda, voedisch2023continual} typically assumed an infinitely sized replay buffer without considering the limited storage capacity on robotic platforms or mobile devices. To address this issue, we use a replay buffer with a fixed maximum size and propose an image diversity-based update mechanism that is comprised of two steps shown in \cref{fig:replay-buffer}. First, determine whether to add the current online data into the replay buffer and, second, if adding the data results in exceeding the predefined buffer size, select a sample that will be removed from the buffer.

Inspired by the loop closure detection in visual SLAM~\cite{li2021deepslam, voedisch2023continual}, we interpret the cosine similarity between image feature maps as a distance measure between two frames $\mathbf{I}_1$ and $\mathbf{I}_2$:
\begin{equation}
    \text{sim}_{\cos} = \cos \left( \text{feat}(\mathbf{I}_1), \text{feat}(\mathbf{I}_2) \right),
\end{equation}
where $\text{feat}(\cdot)$ denotes the respective image features.
In order to determine whether adding a new sample would increase the diversity of the replay buffer, we compute its cosine similarity with respect to all samples that are already in the buffer and take the maximum value.
\begin{equation}
    \text{sim}_\mathbf{B}(\mathbf{I}_t) = \max_{\mathbf{I}_i \in \mathbf{B}} \cos \left( \text{feat}(\mathbf{I}_t), \text{feat}(\mathbf{I}_i) \right),
\end{equation}
where $\mathbf{I}_i \in \mathbf{B}$ refers to the current content of the buffer. If $\text{sim}_\mathbf{B}(\mathbf{I}_t) < \theta_\mathit{th}$, the data triplet associated with $\mathbf{I}_t$ is added to the replay buffer.
In case this results in a buffer size larger than the allowed size, we have to remove a sample from the buffer. Instead of using random sampling, we remove the sample that yields maximal diversity within the remaining samples. Formally, we remove the following sample:
\begin{equation}
    \argmax_{\mathbf{I}_i \in \mathbf{B}} \sum_{\mathbf{I}_j \in \mathbf{B}} \cos \left( \text{feat}(\mathbf{I}_i), \text{feat}(\mathbf{I}_j) \right)
\end{equation}

As described in the next section, we do not update the encoder weights of \net. Therefore, to avoid the overhead of a separate network, we use the encoder of the DepthNet to generate image features.

\subsubsection{Adaptive Visual-Inertial Odometry}
After the replay buffer has been updated, we construct a batch $\mathbf{b}_t$ consisting of the data triplet of the current image $\mathbf{I}_t$ and $N$ samples from the replay buffer.
\begin{equation}
    \mathbf{b}_t = \{ \mathbf{I}_t, \mathbf{I}_1, \mathbf{I}_2, \dots, \mathbf{I}_N \}
\end{equation}
To query the samples from the buffer, we use a uniform probability distribution across all samples and avoid selecting the same sample multiple times if the current size of the buffer is greater than the requested number of samples. To further increase diversity, we augment the replay images in terms of brightness, contrast, saturation, and hue value.
Next, the batch $\mathbf{b}_t$, comprising RGB images and velocity measurements, is fed to the DepthNet to estimate a dense depth map of the center images and to the PoseNet to estimate the camera motion with respect to both neighboring frames. Following the same procedure as during pretraining (see \cref{ssec:ta-network-pretraining}), we then compute the loss $\mathcal{L}$ defined in \cref{eqn:total_loss} and perform backpropagation to update the network weights. Following McCraith~\etal~\cite{mccraith2020monocular}, we do not update the weights of the encoders but only of the decoders.


\subsubsection{Asynchronous \net}
\begin{figure}
    \centering
    \includegraphics[width=.85\linewidth]{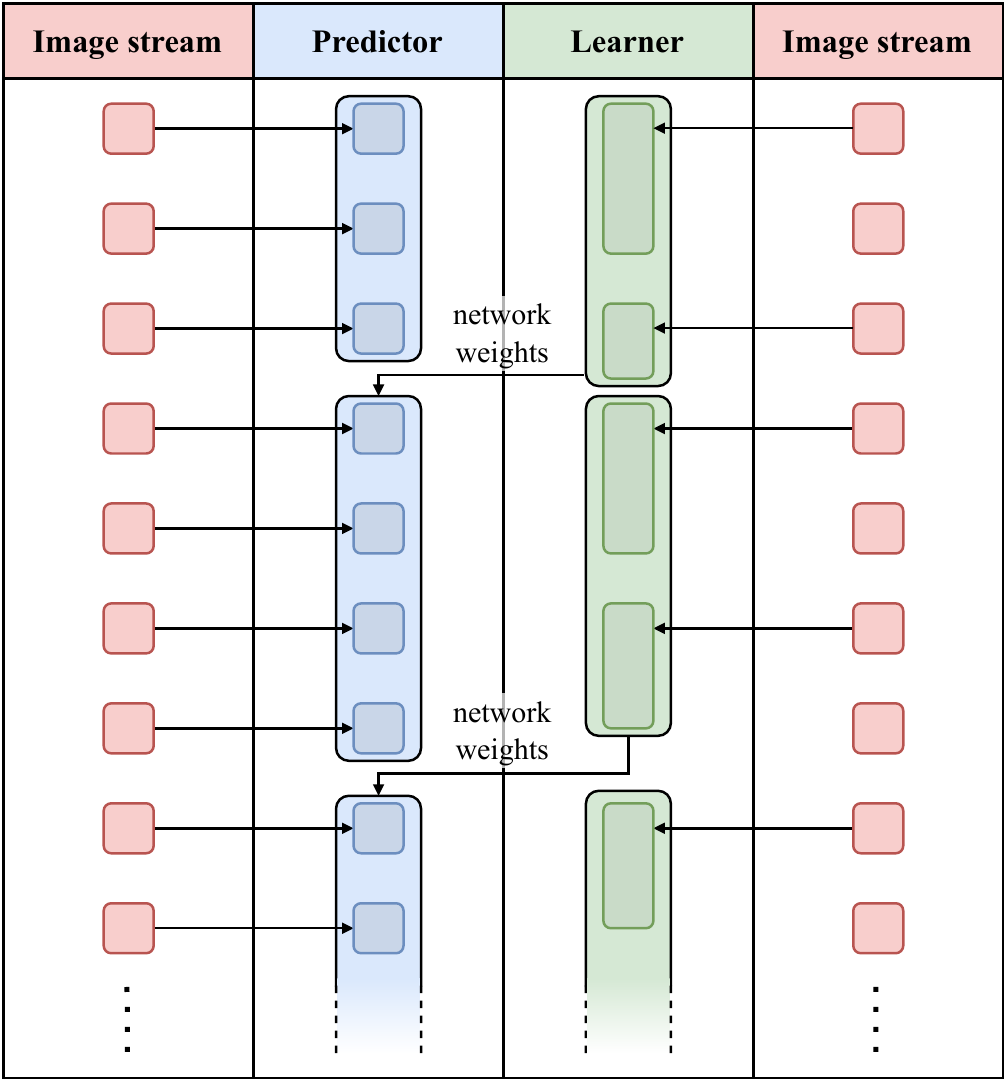}
    \caption{Illustration of the asynchronous variant of \net. While the predictor generates visual odometry estimates in real time, the learner updates the network weights via backpropagation. After a given number of update cycles, the network weights are transferred from the learner to the predictor.}
    \label{fig:asynchronous-version}
    \vspace*{-.3cm}
\end{figure}

Finally, we propose an asynchronous variant of \net to address true continuous inference on robotic platforms in a real-time capable setting. Since multiple update iterations $c$ can result in a situation, in which the network update takes longer than the frame rate of the input camera stream, we also design a version that decouples the VO estimation from the CL updates. As illustrated in \cref{fig:asynchronous-version}, the \textit{predictor} continuously generates VO estimates for each incoming image. The \textit{learner} contains a copy of the network that is updated using the previously introduced online CL strategy but disregards images if the update step takes longer than the time until the next frame is available. Compared to caching frames, this strategy ensures that always the latest information is used to update the network. Then, after a given number of update cycles, the network weights are transferred from the \textit{learner} to the \textit{predictor}.
We include implementations in both ROS and ROS2 in our published code base.

\section{Experimental Evaluation}

In this section, we present extensive experimental results on the efficiency and efficacy of our proposed \net, compared to previous works. We further conduct multiple ablation studies to demonstrate the effect of newly introduced hyperparameters and to justify certain design choices.

Throughout all experiments, we report the translation error $t_\mathit{err}$ (in \%) and the rotation error $r_\mathit{err}$ (in \degree/m) as proposed by Geiger~\etal~\cite{geiger2012are}. These metrics evaluate the error as a function of the trajectory length. To ensure a fair comparison with the base work CL-SLAM~\cite{voedisch2023continual}, we further utilize the set of network weights that is provided by the authors and was pretrained on the Cityscapes Dataset~\cite{cordts2016the}. We also follow CL-SLAM and only consider new frames when the IMU measures a driven distance of at least \SI{0.2}{\meter}.


\subsection{Datasets}

We employ our method on various datasets simulating a diverse set of environments. In particular, we initialize \net with network weights trained on Cityscapes~\cite{cordts2016the} and perform online continual learning on sequences from the KITTI odometry benchmark~\cite{geiger2012are}, the Oxford RobotCar Dataset~\cite{maddern2017the}, and in-house data.


{\parskip=3pt
\noindent\textit{Cityscapes:}
The Cityscapes Dataset~\cite{cordts2016the} is a large-scale autonomous driving dataset that contains RGB images and vehicle metadata such as velocity. It was recorded in 50 cities in Germany, France, and Switzerland. In this work, we use network weights pretrained on Cityscapes that are provided by Vödisch~\etal~\cite{voedisch2023continual}.
}


{\parskip=3pt
\noindent\textit{KITTI:}
The KITTI Dataset~\cite{geiger2012are} is a pioneering autonomous driving dataset that was recorded in Karlsruhe, Germany. For continual learning of new domains, we use images and ground truth poses of multiple sequences from the odometry benchmark and combine them with the respective IMU data from the raw dataset.
}


{\parskip=3pt
\noindent\textit{Oxford RobotCar:}
The Oxford RobotCar Dataset~\cite{maddern2017the} provides multiple recordings of the same route that were captured across one year. We use the included RGB images and the IMU data. To compute the error metrics, we exploit the separately released RTK ground truth positions~\cite{maddern2020real}.
}


{\parskip=3pt
\noindent\textit{In-House:}
Finally, we employ \net on an in-house dataset recorded in Freiburg, Germany. Our robotic platform includes forward-facing RGB cameras and an inertial navigation system (INS), that we use to compute the velocity supervision loss.
}


\subsection{Evaluation of Online Continual Learning}

In this section, we conduct a series of experiments including both simple online domain adaptation from a source~$\mathcal{S}$ to a target domain $\mathcal{T}$ and online continual learning from $\mathcal{S}$ to a sequence of target domains $\{ \mathcal{T}_1, \mathcal{T}_2, \dots \}$. Based on the ablation studies in \cref{ssec:ablation-studies}, we use a replay buffer size of $|\mathbf{B}| = 100$, an update batch size of $|\mathbf{b}_t|=3$, and $c = 5$ backpropagation steps allowing a fair comparison with the base work CL-SLAM~\cite{voedisch2023continual}. We set a similarity threshold of $\theta_\mathit{th} = 0.95$ for the diversity-based update scheme of the replay buffer. For the loss weights, we follow CL-SLAM and use $\gamma = 0.001$ and $\lambda = 0.05$. To compare with other methods, we do not use the asynchronous version but the same learning scheme as in CL-SLAM.


\begin{figure*}
    \centering
    \captionsetup[subfigure]{justification=centering, font=small}
    \subfloat[Seq. 04]{%
        \includegraphics[width=.20\linewidth]{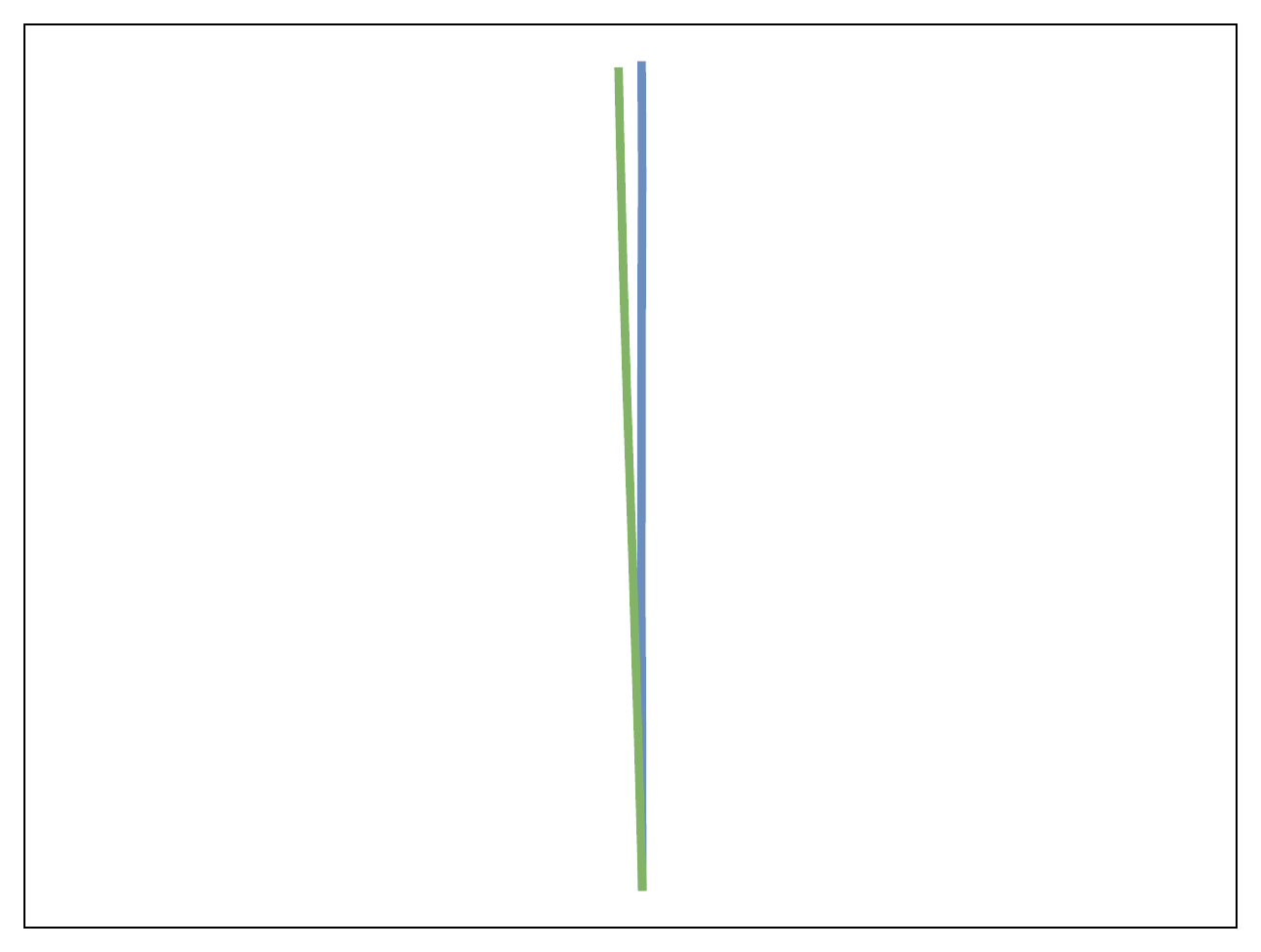}}
    \hfill
    \subfloat[Seq. 05]{%
        \includegraphics[width=.20\linewidth]{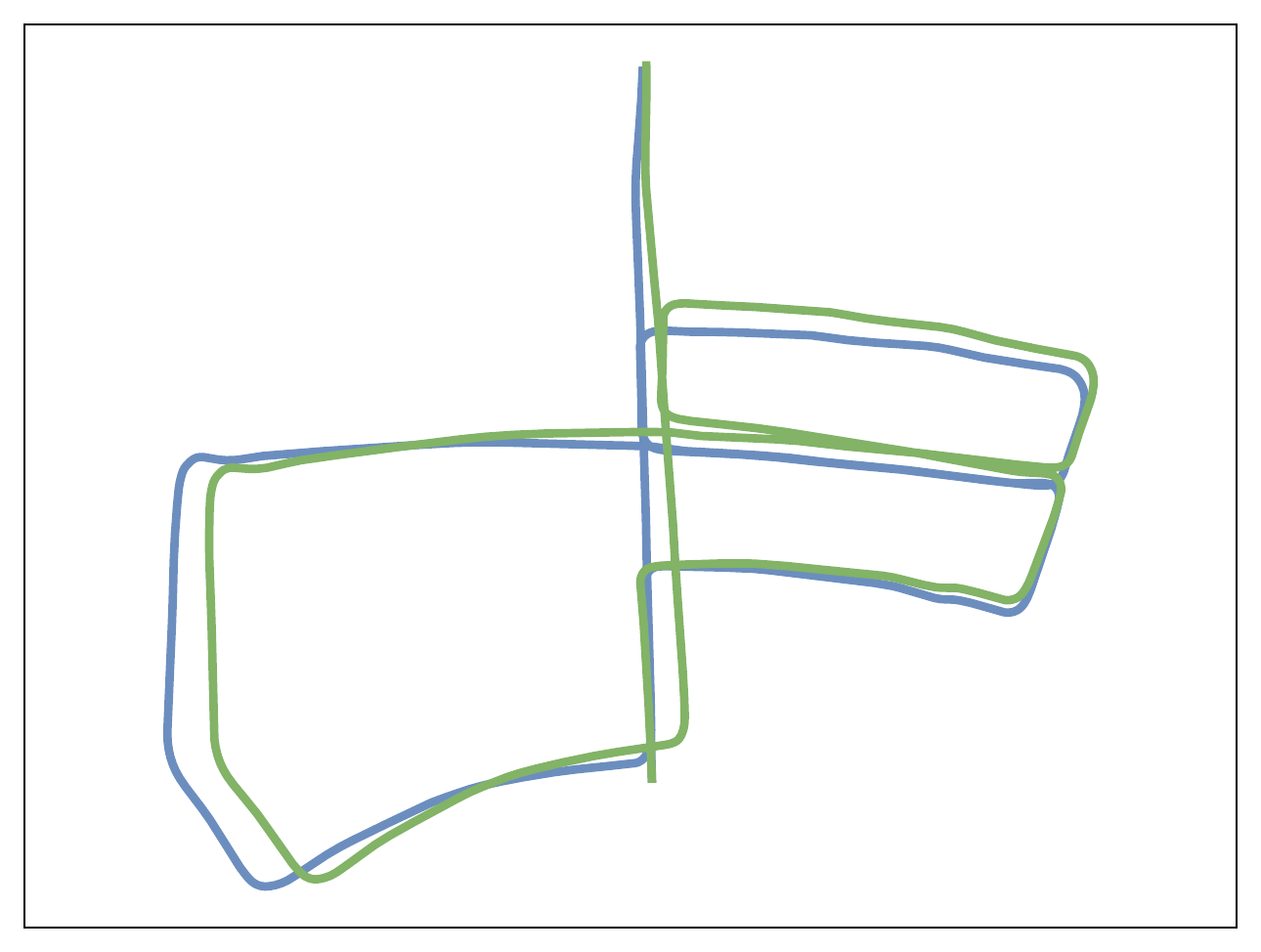}}
    \hfill
    \subfloat[Seq. 06]{%
        \includegraphics[width=.20\linewidth]{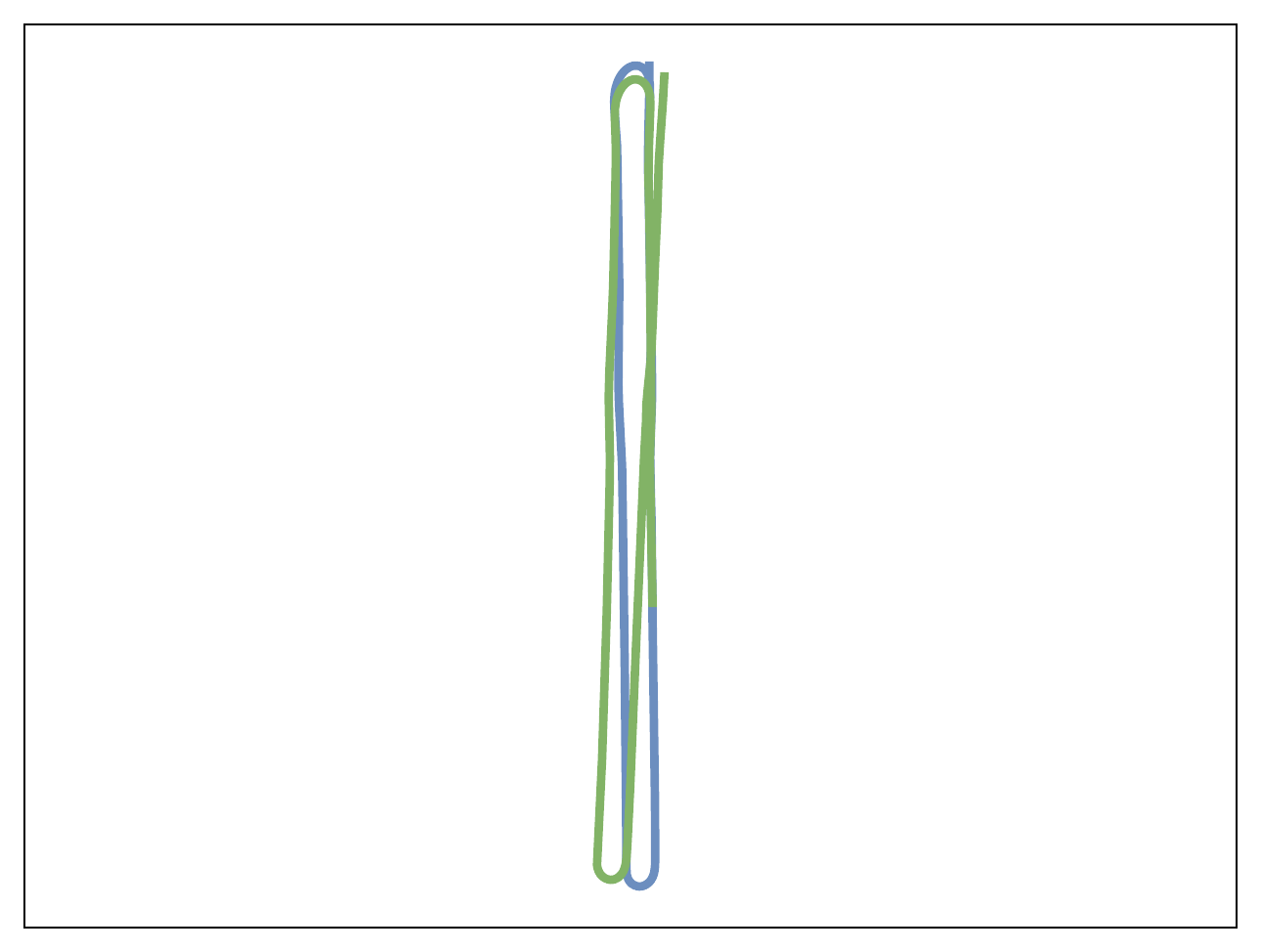}}
    \hfill
    \subfloat[Seq. 07]{%
        \includegraphics[width=.20\linewidth]{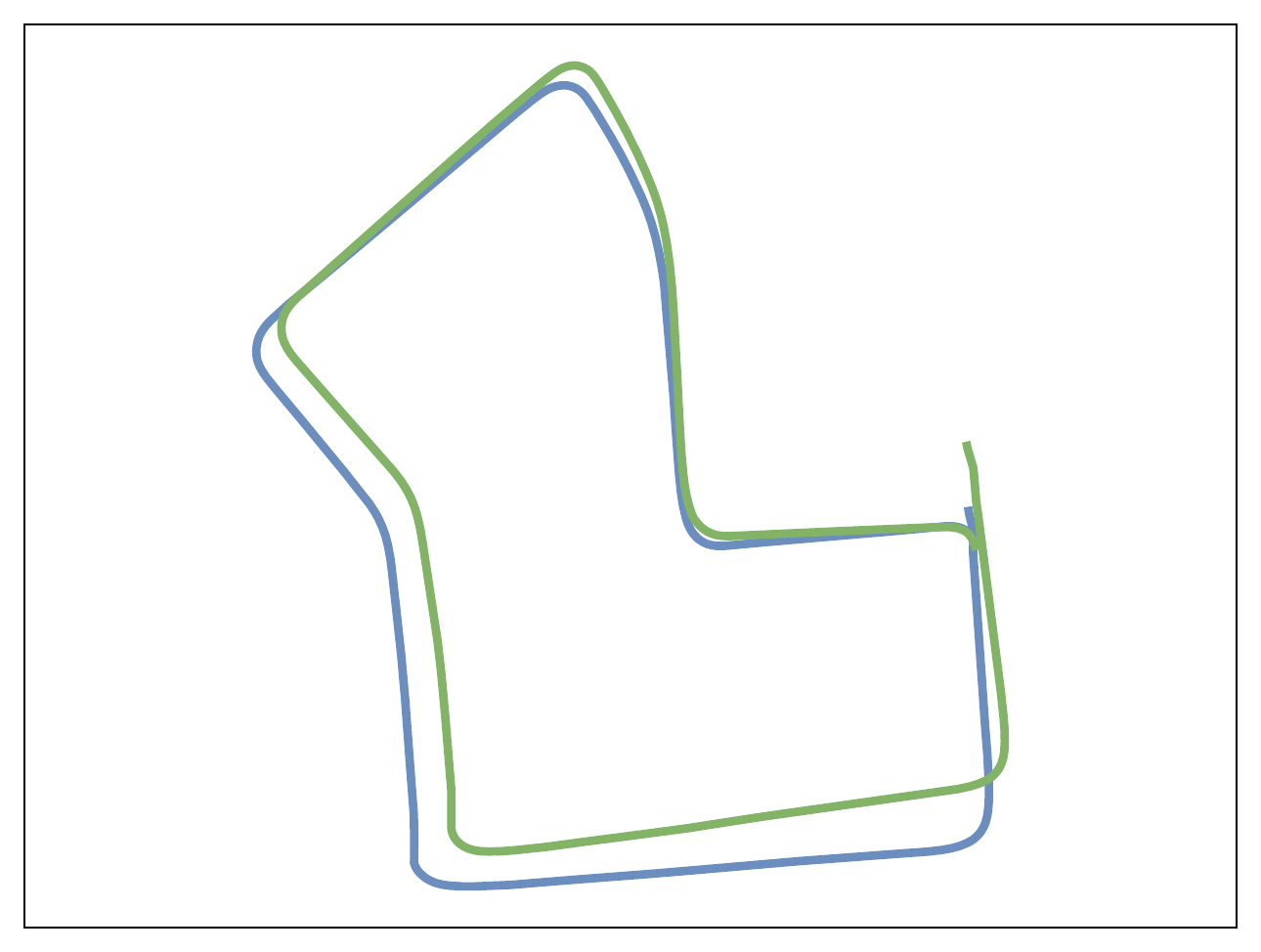}}
    \hfill
    \subfloat[Seq. 10]{%
        \includegraphics[width=.20\linewidth]{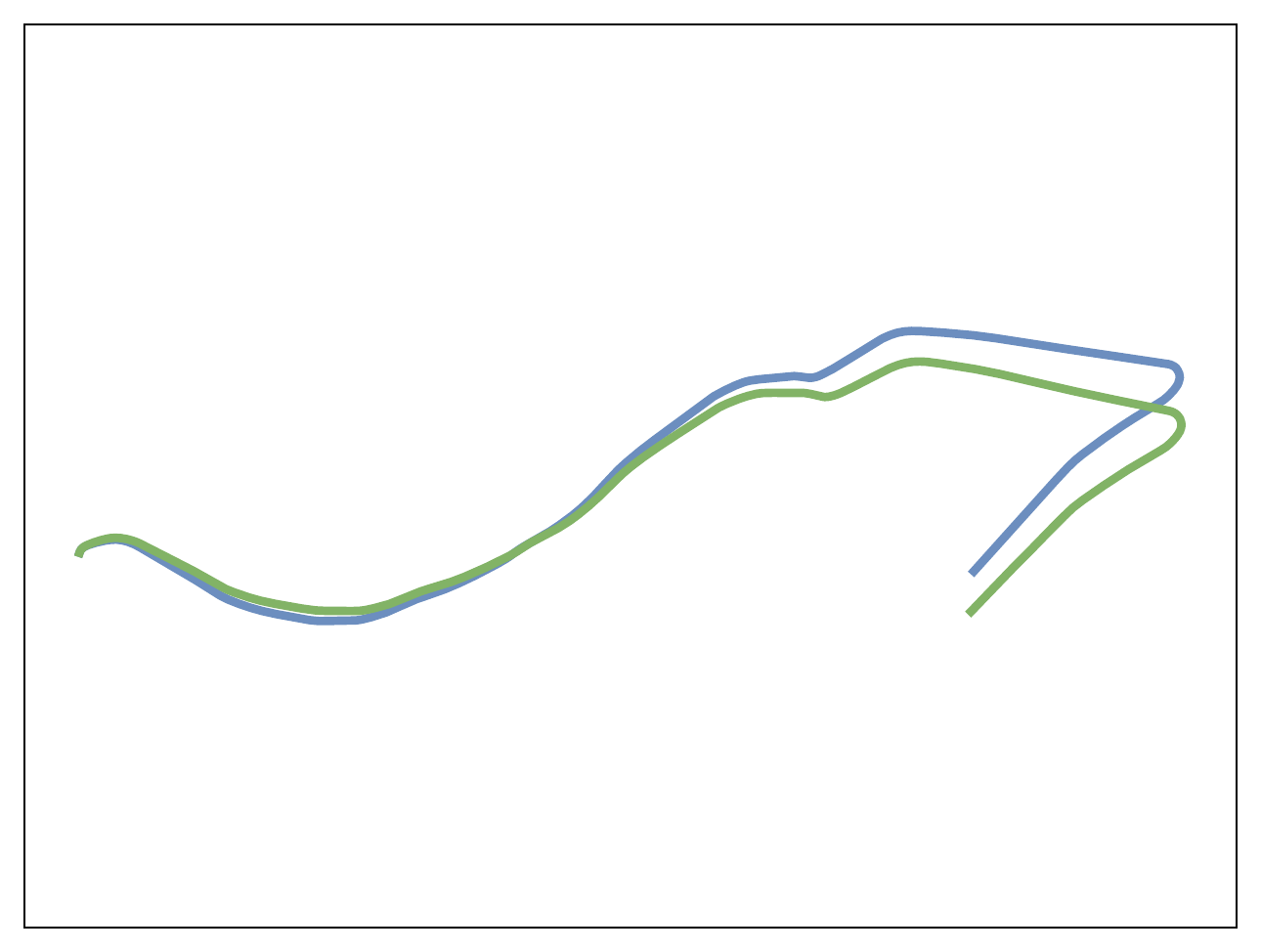}}
    \hfill
    \vspace{0.1cm}
    \includegraphics[width=.25\linewidth]{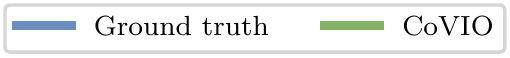}
    \vspace{-0.1cm}
    \caption{Online continual learning results on the KITTI odometry benchmark after pretraining on the Cityscapes dataset.}
    \label{fig:results-kitti}
\end{figure*}

\begin{table*}
\footnotesize
\centering
\caption{Comparison of continual odometry estimation on the KITTI odometry benchmark.}
\label{tab:results-baselines}
\setlength\tabcolsep{11pt}
\begin{threeparttable}
    \begin{tabular}{ l | cc | cc | cc | cc | cc }
         \toprule
         \multirow{2}{*}{Method} & \multicolumn{2}{c|}{Seq. 04} & \multicolumn{2}{c|}{Seq. 05} & \multicolumn{2}{c|}{Seq. 06} & \multicolumn{2}{c|}{Seq. 07} & \multicolumn{2}{c}{Seq. 10} \\
         & t\textsubscript{err} & r\textsubscript{err} & t\textsubscript{err} & r\textsubscript{err} & t\textsubscript{err} & r\textsubscript{err} & t\textsubscript{err} & r\textsubscript{err} & t\textsubscript{err} & r\textsubscript{err} \\
         \midrule
         ORB-SLAM~\cite{murartal2015orbslam} & 0.62 & 0.11 & 2.51 & 0.25 & 7.80 & 0.35 & 1.53 & 0.35 & 2.96 & 0.52 \\
         \midrule
         Only target & 10.72 & 1.69 & 34.55 & 11.88 & 15.20 & 5.62 & 12.77 & 6.80 & 55.27 & 9.50 \\
         DeepSLAM~\cite{li2021deepslam} & 5.22 & 2.27 & \underline{4.04} & 1.40 & 5.99 & 1.54 & 4.88 & 2.14 & \underline{10.77} & 4.45 \\
         \midrule
         Only source & 28.94 & 4.64 & 46.13 & 19.20 & 49.57 & 20.79 & 37.75 & 25.42 & 30.91 & 15.28 \\
         CL-SLAM~\cite{voedisch2023continual} & \underline{4.37} & \textbf{0.51} & 4.30 & \underline{1.01} & \underline{2.53} & \underline{0.63} & \textbf{2.10} & \textbf{0.83} & 11.18 & \underline{1.74} \\
         \net \textit{(ours)} & \textbf{2.11} & \underline{0.53} & \textbf{2.88} & \textbf{0.94} & \textbf{2.13} & \textbf{0.47} & \underline{3.19} & \underline{1.26} & \textbf{3.71} & \textbf{1.55} \\
         \bottomrule
    \end{tabular}
    Comparison of the translation and rotation errors of our \net with baseline methods evaluated on the KITTI odometry benchmark. ``Only target'' and DeepSLAM are trained on sequences \{00, 01, 02, 08, 09\} without further adaptation. ``Only source'', CL-SLAM, and \net are trained on Cityscapes. Both CL-SLAM and \net perform online adaptation on the respective KITTI sequence. 
    The values of CL-SLAM and ``only target'' are reported by Vödisch~\etal~\cite{voedisch2023continual}.
    The errors of the paths predicted by ORB-SLAM are based on ground truth scaling and hence not directly comparable to the other methods.
    The smallest and second smallest errors across the methods producing metric predictions are shown in \textbf{bold} and \underline{underlined}.
\end{threeparttable}
\end{table*}

\subsubsection{Cityscapes to KITTI}
We use our proposed \net to perform online continual adaptation from Cityscapes to KITTI and compare its performance to other methods shown in \cref{tab:results-baselines}. In detail, we compare with the traditional ORB-SLAM~\cite{murartal2015orbslam} as well as the following learning-based methods: ``Only target'' and DeepSLAM~\cite{li2021deepslam} are trained on the KITTI sequences \{00, 01, 02, 08, 09\} without further adaptation; ``only source'', CL-SLAM~\cite{voedisch2023continual}, and \net are trained on Cityscapes with online adaptation to KITTI for both \mbox{CL-SLAM} and \net. Generally, the difference between ``only source'' and ``only target'' demonstrates the domain gap that online adaptation aims to overcome. Our proposed \net outperforms the base method CL-SLAM on the majority of sequences and also improves performance compared to offline training on the target domain. We visualize the predicted and ground truth odometry in \cref{fig:results-kitti}. Note that, unlike CL-SLAM and DeepSLAM, we do not include loop closures in \net.

We further perform online continual learning on all sequences in a sequential manner, \ie, after pretraining on Cityscapes, adapt to sequence 04, then sequence 05, etc., and list the results in \cref{tab:results-continual-learning}. In particular, we compute the translation and rotation errors after each step on all sequences to determine both forward and backward transfer, \ie, the effect on previous and yet unseen future sequences. Since all the sequences of a dataset could be considered to represent similar domains, \eg, the same camera parameters and comparable environments, we observe a general trend of positive forward transfer. Although the performance on previous sequences cannot be fully retained, \net successfully mitigates catastrophic forgetting compared to the initial performance after pretraining on the source domain.

\begin{table*}
\footnotesize
\centering
\caption{Continual odometry estimation results on the KITTI odometry benchmark.}
\label{tab:results-continual-learning}
\begin{threeparttable}
    \begin{tabular}{ c c | cc | cc | cc | cc | cc | cc }
        \toprule
        Sequence & Images & t\textsubscript{err} & r\textsubscript{err} & t\textsubscript{err} & r\textsubscript{err} & t\textsubscript{err} & r\textsubscript{err} & t\textsubscript{err} & r\textsubscript{err} & t\textsubscript{err} & r\textsubscript{err} & t\textsubscript{err} & r\textsubscript{err} \\
        \midrule
        & & \multicolumn{12}{c}{Cityscapes $\xrightarrow{\hspace*{0.8cm}}$ Seq. 04 $\xrightarrow{\hspace*{0.8cm}}$ Seq. 05 $\xrightarrow{\hspace*{0.8cm}}$ Seq. 06 $\xrightarrow{\hspace*{0.8cm}}$ Seq. 07 $\xrightarrow{\hspace*{0.8cm}}$ Seq. 10} \\
        & \\[-1.5ex]
        Seq. 04 & 269 & 28.94 &  4.64 & \cellcolor{Gray}2.11 & \cellcolor{Gray}0.53 &  7.66 &  7.05 &  8.21 &  1.48 &  7.88 &  3.41 &  9.80 & 3.82 \\
        Seq. 05 & 2676 & 46.13 & 19.20 & 59.51 & 16.99 & \cellcolor{Gray}2.85 & \cellcolor{Gray}1.05 &  8.49 &  3.77 &  6.84 &  3.64 & 13.73 & 5.36 \\
        Seq. 06 & 1099 & 49.57 & 20.79 & 65.39 & 22.33 & 20.01 & 10.83 & \cellcolor{Gray}3.08 & \cellcolor{Gray}1.16 &  7.77 &  4.25 &  5.76 & 1.92 \\
        Seq. 07 & 993 & 37.75 & 25.42 & 67.67 & 29.85 &  7.26 &  4.93 &  7.13 &  3.38 & \cellcolor{Gray}6.05 & \cellcolor{Gray}3.53 &  9.60 & 5.33 \\
        Seq. 10 & 1127 & 30.91 & 15.28 & 35.37 & 10.18 & 11.13 &  9.48 &  5.08 &  2.47 & 17.53 &  7.73 & \cellcolor{Gray}2.65 & \cellcolor{Gray}1.15 \\
        \bottomrule
    \end{tabular}
    We continually employ \net on five KITTI sequences after initialization on Cityscapes. The number of images corresponds to the number of update batches of a sequence. The cells highlighted in gray denote the results of the current adaptation step. Along one row, we can measure forward and backward transfer.
\end{threeparttable}
\end{table*}


\subsubsection{Cityscapes to In-House Data}
\begin{figure}
    \centering
    \includegraphics[width=\linewidth]{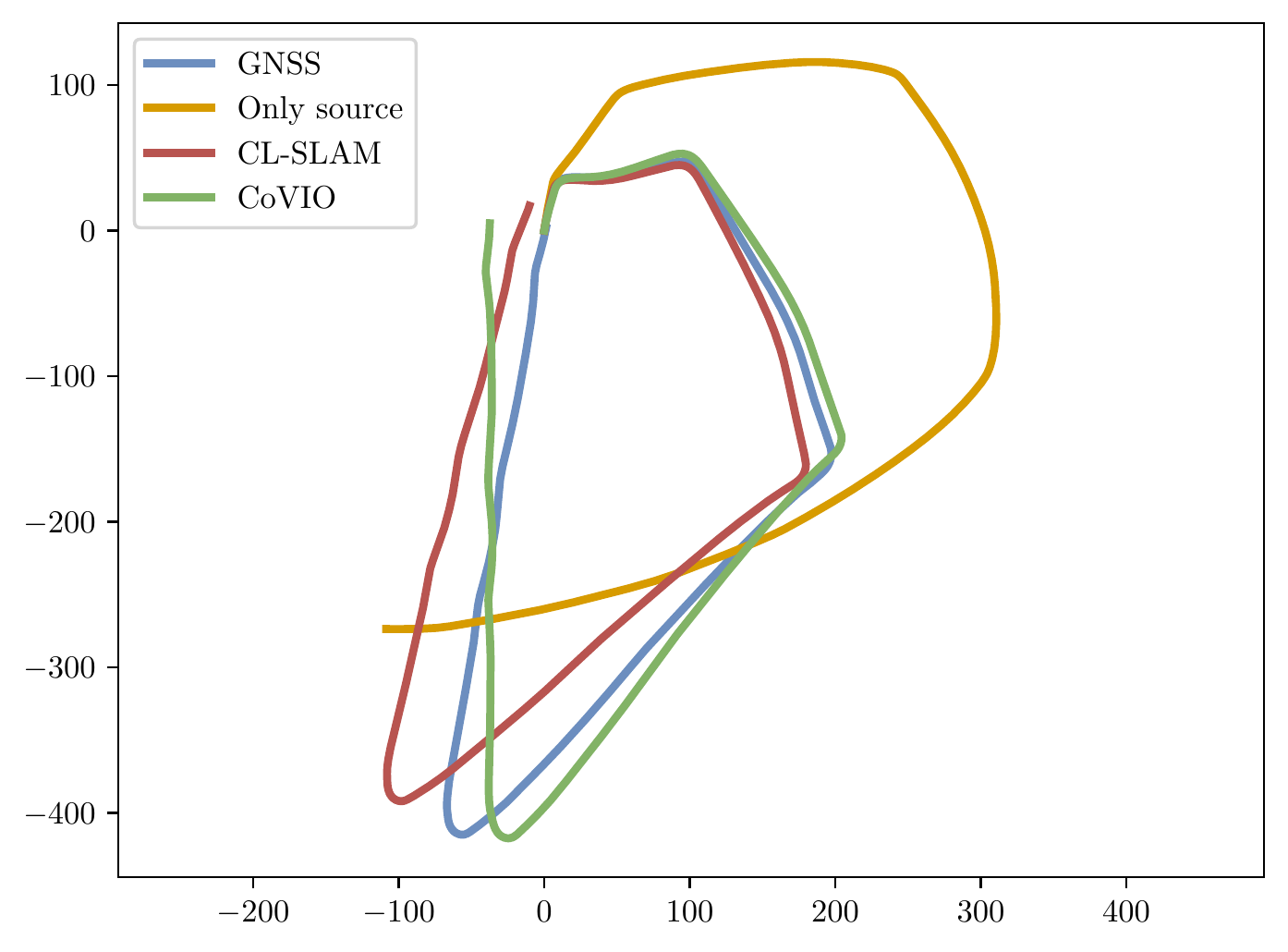}
    \caption{Continual odometry estimation results on in-house data.}
    \label{fig:results-inhouse}
    \vspace*{-.5cm}
\end{figure}

Next, we utilize \net to estimate visual odometry on an in-house dataset, after pretraining on Cityscapes. In \cref{fig:results-inhouse}, we provide a qualitative comparison of \net to CL-SLAM~\cite{voedisch2023continual} with disabled loop closure detection, no online adaptation, and the measured GNSS position. Since we do not have access to highly accurate RTK readings, we omit computing error metrics for this dataset. However, as demonstrated in \cref{fig:results-inhouse}, \net is able to maintain accurate odometry tracking for a longer distance than CL-SLAM.

\subsubsection{Cityscapes to KITTI and RobotCar}
We further investigate the capability of \net to retain knowledge in a multi-target setting. In detail, we perform the same experiment as conducted by CL-SLAM~\cite{voedisch2023continual}. After initialization on Cityscapes, we sequentially deploy \net to KITTI sequence 09, a sequence from RobotCar, KITTI sequence 10, and another sequence from RobotCar. For further details on the RobotCar sequences, we refer the reader to \cite{voedisch2023continual}. In \cref{tab:adaptation-retention-quality}, we report the adaptation quality (AQ) and the retention quality (RQ) as introduced by Vödisch~\etal~\cite{voedisch2023continual}. Broadly, the AQ score measures the ability of a method to adapt to a previously unseen environment, whereas the RQ measures the ability to retain long-term knowledge when being redeployed to a previously seen domain. Compared to CL-SLAM, \net improves the AQ score and, with a high margin, the RQ with respect to the translation error. Although RQ\textsubscript{rot} suffers from a small decrease, the absolute rotation errors on the four considered sequences are smaller than those of CL-SLAM, hence smaller differences between with and without an intermediate domain influence the RQ more strongly.

\begin{table}
\footnotesize
\centering
\caption{Comparison of adaptation and retention quality.}
\label{tab:adaptation-retention-quality}
\setlength\tabcolsep{4pt}
\begin{threeparttable}
    \begin{tabular}{ cc | cc cc }
        \toprule
        Previous & Current & \multicolumn{2}{c}{CL-SLAM~\cite{voedisch2023continual}} & \multicolumn{2}{c}{\net \textit{(ours)}} \\
        sequences & sequence & t\textsubscript{err} & r\textsubscript{err} & t\textsubscript{err} & r\textsubscript{err} \\
        \midrule
        $c_t$ & $k_9$ & \textbf{2.50} & \textbf{0.37} & 3.89 & 1.49 \\
        $c_t$ & $r_1$ & 28.94 & 5.63 & \textbf{6.62} & \textbf{2.61} \\
        $c_t \shortto r_1$ & $k_9$ & \textbf{3.24} & \textbf{0.54} & 4.09 & 1.18 \\
        $c_t \shortto k_9$ & $r_1$ & 30.13 & 5.87 & \textbf{11.00} & \textbf{3.44} \\
        \midrule
        \multicolumn{2}{c|}{$\Rightarrow$ AQ\textsubscript{trans} / AQ\textsubscript{rot}} & 0.85 & 0.98 & \textbf{0.94} & \textbf{0.99} \\
        \midrule
        \midrule
        $c_t \shortto k_9 \shortto r_1$ & $k_{10}$ & 4.85 & 1.59 & \textbf{1.86} & \textbf{0.70} \\
        $c_t \shortto k_9 \shortto r_1 \shortto k_{10}$ & $r_2$ & 20.50 & 4.77 & \textbf{5.66} & \textbf{3.99} \\
        $c_t \shortto k_9$ & $k_{10}$ & 7.48 & 1.63 & \textbf{1.43} & \textbf{0.73} \\
        $c_t \shortto k_9 \shortto r_1$ & $r_2$ & 16.41 & 4.58 & \textbf{7.67} & \textbf{3.42} \\
        \midrule
        \multicolumn{2}{c|}{$\Rightarrow$ RQ\textsubscript{trans} / RQ\textsubscript{rot} $\times 10^{-3}$} & -7.30 & \textbf{-0.42} & \textbf{7.89} & -1.53 \\
        \bottomrule
    \end{tabular}
    Adaptation quality (AQ) and retention quality (RQ)~\cite{voedisch2023continual} with respect to the translation and rotation errors.
    $c_t$ denotes the Cityscapes training set, $k_9$ and $k_{10}$ refer to sequences 09 and 10 of the KITTI odometry benchmark, and $r_1$ and $r_2$ correspond to sequences~\cite{voedisch2023continual} from RobotCar.
    The values of CL-SLAM are reported by the authors.
    The best scores in each category are shown in \textbf{bold}.
\end{threeparttable}
\end{table}


\subsection{Ablation Study}
\label{ssec:ablation-studies}

In this section, we present the results of various ablation studies substantiating the design choices for the sizes of the update batch $\mathbf{b}_t$ and the replay buffer $\mathbf{B}$. We further demonstrate that \net is less sensitive to the number of backpropagation steps per update batch than a previous method. In the following studies, we always report the translation and rotation errors of sequences 04 and 06.


\subsubsection{Size of the Update Batch}
We first investigate the effect of varying sizes of the update batch, \ie, the number of replay samples. In \cref{tab:ablation-batch-size}, we list the errors for batch sizes $\mathbf{b}_t = \{ 1, 2, 3, 4, 5 \}$ given an unlimited replay buffer. Note that $\mathbf{b} = 1$ implies that experience replay is disabled. Therefore, this strategy corresponds to CL-SLAM~\cite{voedisch2023continual} for source-to-target domain adaptation. Generally, the translation error can be reduced by using replay data, whereas the rotation error is the smallest when only training with the current sample. As we deem the translation error more important in terms of mapping and localization accuracy, we select $\mathbf{b} = 3$.

\begin{table}[t]
\footnotesize
\centering
\caption{Ablation study on the size of the update batch.}
\label{tab:ablation-batch-size}
\setlength\tabcolsep{9pt}
\begin{threeparttable}
    \begin{tabular}{c | cc | cc }
        \toprule
        \multirow{2}{*}{Batch size} & \multicolumn{2}{c|}{Seq. 04} & \multicolumn{2}{c}{Seq. 06} \\
        & t\textsubscript{err} & r\textsubscript{err} & t\textsubscript{err} & r\textsubscript{err} \\
        \midrule
        1 & 3.56 & \textbf{0.15} & 2.30 & \textbf{0.18} \\
        2 & 2.97 & 0.59 & \textbf{1.81} & \underline{0.50} \\
        \rowcolor{Gray}
        3 & \textbf{2.79} & \underline{0.54} & \underline{1.98} & 0.59 \\
        4 & \underline{2.89} & 0.73 & 1.99 & 0.54 \\
        5 & \underline{2.89} & 0.63 & 2.46 & 0.70 \\
        \bottomrule
    \end{tabular}
    In this study, we use a replay buffer of infinite size.
    Batch sizes greater than 1 imply using replay data in addition to the online image, \ie, the first row corresponds to the strategy of CL-SLAM.
    The smallest and second smallest errors are shown in \textbf{bold} and \underline{underlined}, respectively.
\end{threeparttable}
\end{table}


\subsubsection{Size of the Replay Buffer}
In the next study, we restrict the size of the replay buffer to address both scalability of the method and the limited storage capacity on mobile devices and robotic platforms. In detail, we report the error for buffer sizes $|\mathbf{B}| = \{ 10, 25, 50, 100, \infty \}$ in \cref{tab:ablation-buffer-size}. For all the buffers of limited size, we both enable and disable our proposed diversity-based updating mechanism. Interestingly, the positive effect of enforcing a high diversity is more pronounced for sequence 04, where \net generally yields smaller translation errors with fewer samples in the buffer. It should further be noted that due to the length of the sequences, the diversity-based buffer contains the same samples for $|\mathbf{B}_\text{Seq. 04}| = \{ 25, 50, 100\}$ and $|\text{B}_\text{Seq. 06}| = \{ 50, 100\}$. For \net, we select $|\mathbf{B}| = 100$ to account for the increased storage requirements in a multi-target setting.

\begin{table}
\footnotesize
\centering
\caption{Ablation study on the size of replay buffer.}
\label{tab:ablation-buffer-size}
\begin{threeparttable}
    \begin{tabular}{ c | c | cc | cc }
        \toprule
        Buffer & Diversity & \multicolumn{2}{c|}{Seq. 04} & \multicolumn{2}{c}{Seq. 06} \\
        size & update & t\textsubscript{err} & r\textsubscript{err} & t\textsubscript{err} & r\textsubscript{err} \\
        \midrule
        $\infty$ &   & 2.79 & 0.54 & 1.98 & 0.59 \\
        \midrule
        100 &        & 2.62 & 0.52 & \textbf{1.75} & \underline{0.48} \\
        \rowcolor{Gray}
        100 & \cmark & \textbf{2.11} & 0.53 & 2.13 & \textbf{0.47} \\
        50 &         & 2.64 & 0.42 & 2.72 & 0.91 \\
        50 & \cmark  & \textbf{2.11} & 0.53 & 2.13 & \textbf{0.47} \\
        25 &         & 2.51 & \underline{0.40} & 2.42 & 0.77 \\
        25 & \cmark  & \textbf{2.11} & 0.53 & 2.20 & 0.50 \\
        10 &         & 2.82 & \textbf{0.33} & \underline{2.08} & 0.64 \\
        10 & \cmark  & \underline{2.12} & 0.56 & 2.21 & 0.59 \\
        \bottomrule
    \end{tabular}
    In this study, we use a batch size of 3.
    The effectively used buffer size of sequence 04 is the same for 25, 50, and 100. Similarly, in sequence 06 the same number of samples is added when buffer sizes of 50 and 100 are available.
    The first row corresponds to the strategy of CL-SLAM.
    The smallest and second smallest errors are shown in \textbf{bold} and \underline{underlined}, respectively.
\end{threeparttable}
\end{table}


\begin{table}
\footnotesize
\centering
\caption{Ablation study on the number of update cycles.}
\label{tab:ablation-update-cycles}
\setlength\tabcolsep{9.5pt}
\begin{threeparttable}
    \begin{tabular}{ c | cc | cc }
        \toprule
        \multirow{2}{*}{Update cycles} & \multicolumn{2}{c|}{Seq. 04} & \multicolumn{2}{c}{Seq. 06} \\
        & t\textsubscript{err} & r\textsubscript{err} & t\textsubscript{err} & r\textsubscript{err} \\
        \midrule
        1 & 3.13 & 1.65 & 4.07 & 1.14 \\
        2 & 2.80 & 1.06 & 2.62 & 0.64 \\
        3 & 2.49 & 0.77 & 2.61 & 0.63 \\
        4 & 2.29 & 0.61 & 2.31 & 0.52 \\
        \rowcolor{Gray}
        5 & \underline{2.11} & \underline{0.53} & \textbf{2.13} & \textbf{0.47} \\
        6 & \textbf{1.98} & \textbf{0.50} & \underline{2.15} & \underline{0.49} \\
        \bottomrule
    \end{tabular}
    For a fair comparison, we use the same number of update cycles $c=5$ as CL-SLAM~\cite{voedisch2023continual}.
    The smallest and second smallest errors are shown in \textbf{bold} and \underline{underlined}, respectively.
\end{threeparttable}
\end{table}

\subsubsection{Number of Update Cycles}
Lastly, we report the sensitivity of \net with respect to the number of backpropagation steps $c$ for a single update batch. As shown in \cref{tab:ablation-update-cycles}, more steps decrease both the errors confirming the results of CL-SLAM~\cite{voedisch2023continual}. However, in contrast to the performance reported for CL-SLAM, \net is noticeably less sensitive and already yields relatively small errors for $c = 1$. We conclude that this is caused by using experience replay also for the online learner. To enable direct comparison with CL-SLAM, we use the same value $c = 5$.

\section{Conclusion}
In this paper, we presented \net for online continual learning of visual-inertial odometry. \net exploits an unsupervised training scheme during inference time and thus seamlessly adapts to new domains. In particular, we explicitly address the shortcomings of previous works by designing a lightweight network architecture and by proposing a novel fixed-size replay buffer that maximizes image diversity. Using experience replay, \net successfully mitigates catastrophic forgetting while achieving efficient online domain adaptation. We further provide an asynchronous version of \net separating the core motion estimation from the network update step, hence allowing true continuous inference in real time. In extensive evaluations, we demonstrated that \net outperforms competitive baselines. We also made the code and models publicly available to facilitate future research. Future work will focus on extending this work to a multi-task setup for robotic vision systems.

\section*{Acknowledgment}
This work was funded by the European Union’s Horizon 2020 research and innovation program under grant agreement No 871449-OpenDR.
We thank Kürsat Petek for the extensive discussions concerning the replay buffer and for supporting the collection of our in-house data.
We further thank Ahmet Selim Çanakçı for helping us with the implementation of the asynchronous version of \net.


{\small
\bibliographystyle{ieee_fullname}
\bibliography{references}
}

\end{document}